\documentclass{article}

\usepackage[preprint]{neurips_2026}

\AtBeginDocument{%
  \newgeometry{
    top=2.5cm,
    bottom=2.5cm,
    left=2.5cm,
    right=2.5cm,
    headheight=14pt
  }%
}

\usepackage{fancyhdr}
\fancyheadoffset[R]{0pt}

\fancypagestyle{firstpage}{%
  \fancyhf{}%
  \rhead{\today}%
}
\fancypagestyle{plain}{%
  \fancyhf{}%
  \rhead{\today}%
}
\makeatletter
\renewcommand{\@toptitlebar}{}
\renewcommand{\@bottomtitlebar}{}
\makeatother

\usepackage{titletoc}
\usepackage[utf8]{inputenc}
\usepackage[T1]{fontenc}
\usepackage{hyperref}
\usepackage{url}
\usepackage{booktabs}
\usepackage{amsfonts}
\usepackage{nicefrac}
\usepackage{microtype}
\usepackage{mathrsfs}
\usepackage{graphicx}
\usepackage{subfigure}
\usepackage{multirow}
\usepackage{amsmath}
\usepackage{float}
\usepackage{amsthm}
\usepackage{comment}
\usepackage{multicol}
\usepackage{appendix}
\usepackage{amssymb}
\usepackage{wrapfig}
\usepackage{bbm}
\usepackage{bbding}
\usepackage{framed}
\usepackage{setspace}
\usepackage{pifont}
\usepackage[table]{xcolor}
\usepackage{makecell}
\usepackage[linesnumbered,ruled,lined]{algorithm2e}
\usepackage{pythonhighlight}
\usepackage{tcolorbox}
\usepackage{enumitem}
\usepackage{xspace}
\usepackage{textcomp}
\usepackage{fontawesome5}
\definecolor{HFOrange}{HTML}{FF9D00}
\definecolor{GitHubGray}{HTML}{24292E}

\newcommand{\ours}[0]{\texttt{TEMPO}\xspace}

\definecolor{mycolor1}{rgb}{0.82,0.70,0.54}
\definecolor{mycolor2}{rgb}{0.0,0.51,0.22}
\definecolor{mycolor3}{rgb}{0.80, 0.48, 0.37}
\definecolor{mycolor4}{rgb}{0.02, 0.33, 0.68}
\definecolor{mycolor5}{rgb}{0.86, 0.11, 0.11}
\definecolor{NavyBlue}{RGB}{30,50,150}
\definecolor{Salmon}{HTML}{FA8072}
\definecolor{lightblue}{RGB}{220,230,245}
\definecolor{darkblue}{RGB}{0,51,102}

\title{TEMPO: Scaling Test-time Training for Large Reasoning Models}

\author{
\\
Qingyang~Zhang\textsuperscript{1,4}\thanks{Equal contribution.},
Xinke Kong\textsuperscript{1}\addtocounter{footnote}{-1}\footnotemark,
Haitao Wu\textsuperscript{1},
Qinghua Hu\textsuperscript{1},
Minghao Wu\textsuperscript{2},
Baosong Yang\textsuperscript{2},
\\
Yu Cheng\textsuperscript{3},
Yun Luo\textsuperscript{4}\thanks{Co-supervised. Correspondence to Yun Luo, Ganqu Cui and Changqing Zhang.},
Ganqu Cui\textsuperscript{4}\addtocounter{footnote}{-1}\footnotemark,
Changqing Zhang\textsuperscript{1}\addtocounter{footnote}{-1}\footnotemark
\\
Tianjin University\textsuperscript{1}
Tongyi Lab, Alibaba Group\textsuperscript{2}
The Chinese University of Hong Kong\textsuperscript{3}
Shanghai AI Lab\textsuperscript{4}
}

\begin{document}

\maketitle

\begin{center}
\vspace{-2em}
\small
\href{https://qingyangzhang.github.io/tempo-homepage}{{\color{NavyBlue}\faGlobe}\;\textcolor{NavyBlue}{\textbf{Project Page}}}%
\quad\textcolor{gray!60}{|}\quad
\href{https://github.com/QingyangZhang/TEMPO}{{\color{GitHubGray}\faGithub}\;\textcolor{GitHubGray}{\textbf{GitHub}}}%
\quad\textcolor{gray!60}{|}\quad
\href{https://huggingface.co/collections/qingyangzhang/tempo}{{\color{HFOrange}\faSmileBeam}\;\textcolor{HFOrange}{\textbf{HuggingFace}}}%
\end{center}

\thispagestyle{firstpage}

\begin{abstract}
Test-time training (TTT) adapts model parameters on unlabeled test instances during inference time, which continuously extends capabilities beyond the reach of offline training. Despite initial gains, existing TTT methods for LRMs plateau quickly and do not benefit from additional test-time compute. Without external calibration, the self-generated reward signal increasingly drifts as the policy model evolves, leading to both performance plateaus and diversity collapse. We propose \ours, a TTT framework that interleaves policy refinement on unlabeled questions with periodic critic recalibration on a labeled dataset. By formalizing this alternating procedure through the Expectation-Maximization (EM) algorithm, we reveal that prior methods can be interpreted as incomplete variants that omit the crucial recalibration step. Reintroducing this step tightens the evidence lower bound (ELBO) and enables sustained improvement. Across diverse model families (Qwen3 and OLMO3) and reasoning tasks, \ours improves OLMO3-7B on AIME 2024 from 33.0\% to 51.1\% and Qwen3-14B from 42.3\% to 65.8\%, while maintaining high diversity. Code is available at \href{https://github.com/QingyangZhang/TEMPO}{this url}.
\end{abstract}

\begin{figure}[H]
    \includegraphics[width=0.99\textwidth]{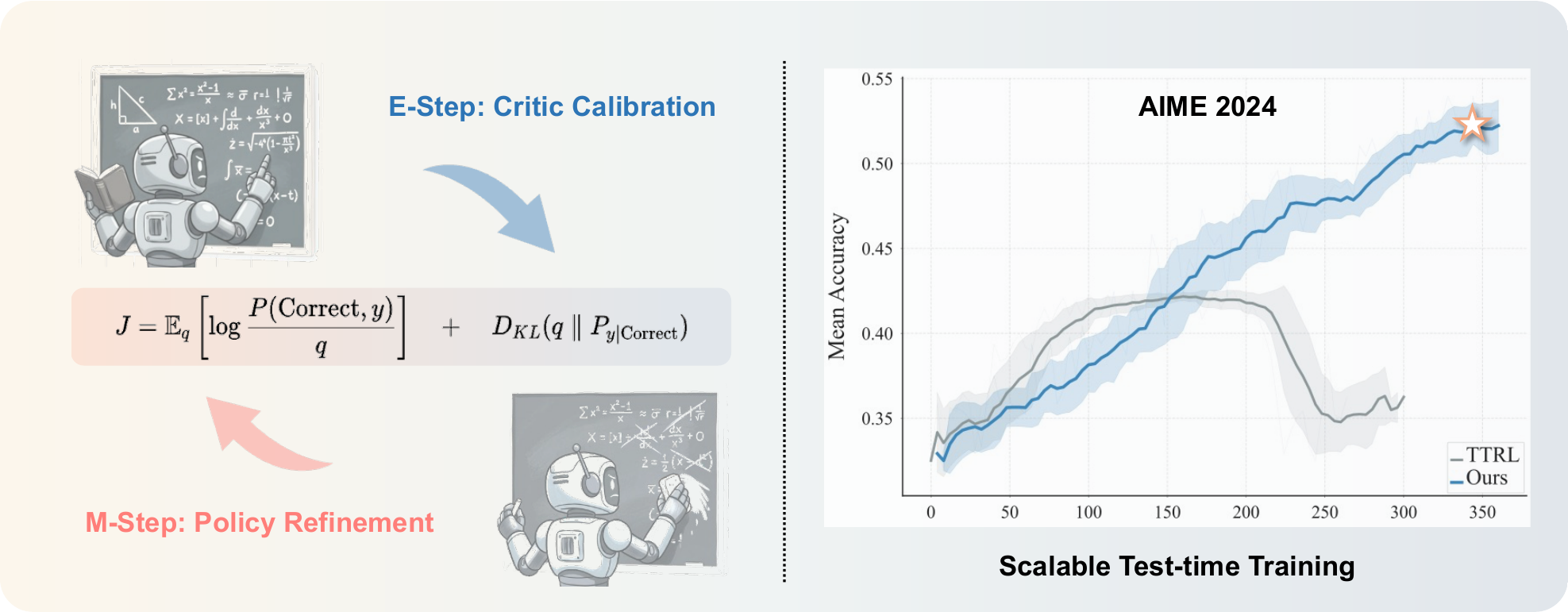}
    \caption{\textbf{Scalability of \ours on the AIME benchmark.} \ours alternates between an E-Step (critic recalibration on labeled data) and an M-Step (policy refinement on unlabeled test questions), guided by a critic that provides quality-aware scores. Representative self-rewarding TTT baselines such as TTRL (grey curves) plateau and collapse after initial gains. In contrast, \ours (blue curve) sustains a consistent upward trajectory over 350 steps by periodically grounding the critic in external supervision, demonstrating that additional test-time compute translates into scalable performance gains on complex open problems.}
    \label{fig:framework}
\end{figure}

\section{Introduction}\label{sec:intro}

Large reasoning models (LRMs) have demonstrated remarkable capabilities on complex reasoning tasks, including logic puzzle-solving and Olympiad-level mathematics and physics~\cite{cui2025process,chen2025p1,luo2026p1}. These models achieve their performance by allocating extensive computation at test time through extended reasoning chains. However, their parameters remain static after training, which prevents them from incorporating knowledge acquired during test-time experience. Test-time training (TTT) addresses this limitation by enabling models to update their parameters on unlabeled test data, thereby extending their reasoning capabilities beyond the original training distribution. Recent methods such as EMPO~\cite{zhang2025right}, TTRL~\cite{zuo2025ttrl}, and Theta-Evolve~\cite{wang2025thetaevolve} employ self-generated reward signals such as entropy, majority voting, or self-consistency to refine reasoning policies via reinforcement learning without ground-truth labels. Practical implementations such as Cursor's real-time reinforcement learning for Composer demonstrate the efficacy of test-time training in dynamically adapting model to complex, interactive coding environments~\cite{cursor2026realtime}.

Despite promising initial results, existing TTT methods for LRMs exhibit two fundamental limitations. First, they rely on heuristic reward signals that are intrinsically bounded by the model's initial capabilities, leading to performance plateaus as self-improvement saturates~\cite{zuo2026how}. Second, these methods tend to collapse output diversity in pursuit of higher average performance, ultimately degrading reasoning quality~\cite{zhang2025freelunchrethinkinginternal}. Both issues share a common root cause: there are no ground-truth labels available at test time, and the reward signal must be inferred from the model's own outputs. As the model becomes increasingly confident in a narrow set of reasoning patterns, these heuristic signals systematically overestimate the quality of self-generated responses, creating a self-reinforcing loop that drives both saturation and diversity collapse.

To this end, we propose Test-time Expectation-Maximization Policy Optimization (\ours), a TTT framework that decouples reward generation from policy optimization through an alternating actor-critic design (Figure~\ref{fig:framework}). \ours operates in two stages: (i) \textit{\textit{Policy Refinement}}, where the actor generates reasoning trajectories on unlabeled test prompts and optimizes against rewards from a critic model, enabling on-the-fly adaptation to novel problems; and (ii) \textit{\textit{Critic Recalibration}}, where the critic is periodically updated using verifiable rewards from a labeled dataset. By maintaining a grounded critic, \ours provides a stable training signal that avoids the reward drift responsible for prior failures. We formalize this alternating procedure through the Expectation-Maximization (EM) algorithm. The key insight is that response correctness is an unobserved latent variable at test time. Thus, optimizing the policy without ground-truth labels is naturally framed as maximizing a lower bound on the expected reward. In this view, the critic recalibration corresponds to the E-step (estimating the posterior distribution over correct responses), while policy optimization corresponds to the M-step (updating model parameters given those estimates). This framing reveals that existing self-rewarding methods are degenerate variants that execute only the M-step, causing the estimated posterior to drift from true correctness. Restoring the E-step through periodic calibration tightens the lower bound and sustains improvement over extended training horizons.

Experimental results validate both the effectiveness and scalability of \ours. On AIME 2024 and 2025, \ours improves OLMO3-7B from 33.0\% and 26.3\% to 51.1\% and 37.0\%. For Qwen3-14B,  \ours pushes the accuracy from 42.3\% and 37.1\% to 65.8\% and 44.6\%, achieves an absolute gain of 23.5 and 7.5 percentage points, respectively. More importantly, \ours maintains high pass@K scores where baselines suffer from diversity collapse. Beyond mathematics, \ours generalizes to non-math reasoning domains, including logic puzzles and STEM tasks, confirming that the alternating critic-policy design is not domain-specific.

Our contributions are summarized as follows:

\begin{itemize}\item We propose \ours, a test-time training framework that achieves sustained performance gains through an alternating actor-critic optimization, avoiding the diversity collapse and performance plateaus of prior LRMs self-training methods (Sec. \ref{sec:method}).

\item We provide a unified analysis that characterizes existing TTT methods as incomplete EM procedures that omit the crucial posterior recalibration. By identifying the missing E-step as the root cause of scalability failures, this perspective yields a principled remedy: restoring periodic critic calibration on labeled data (Sec. \ref{sec:discussion}).

\item We conduct extensive experiments across model families, scales (OLMO3-7B, Qwen3-8B, and Qwen3-14B), and five reasoning benchmarks spanning math, logic puzzles, and STEM. \ours demonstrates both superior accuracy and preserved output diversity, confirming that the alternating design generalizes beyond mathematical reasoning (Sec. \ref{sec:experiments}).\end{itemize}

\section{Related Work}\label{sec:related_work}
This section surveys two lines of prior work that motivate our approach: self-rewarding RL methods that avoid ground-truth labels but suffer from reward self-reinforcement, and existing TTT methods for reasoning models that share the structural deficiency of omitting reward calibration.

\paragraph{Self-rewarding reinforcement learning.}
RLVR, first formalized by Tulu-V3~\cite{lambert2024tulu3}, has emerged as the dominant paradigm for incentivizing reasoning capability in LLMs~\cite{guo2025deepseek, shao2024deepseekmath, cui2025process}, but its reliance on labeled data motivates self-rewarding alternatives. Previous works leverage intrinsic rewards such as entropy~\cite{zhang2025right, gao2025one}, self-certainty~\cite{zhao2025learning}, or reasoning topology~\cite{wang2026sarl} for self-training, without dependency on external supervision. For example, SARL~\cite{wang2026sarl} constructs rewards from the graph structure of intermediate reasoning steps, optimizing to encourage locally coherent and efficient thinking. LaSeR~\cite{yang2025laser} demonstrates that the logit of the last token can serve as an effective self-rewarding signal, achieving superior reasoning accuracy and inference-time scaling with only one additional token of computation. However, these methods tend to collapse the output distribution and plateau as the reward signal becomes self-reinforcing~\cite{zhang2025freelunchrethinkinginternal, zuo2026how}. Our approach avoids this by decoupling reward generation (a critic periodically re-calibrated on labeled data) from policy optimization (on unlabeled test data), preventing the self-reinforcement loop.

\paragraph{Test-time training for reasoning models.}
TTT originated in computer vision, where models continuously update their parameters on each test instance at inference time to fill the gap of distribution shifts~\cite{grandvalet2004semi, wang2021tent, zhang2024come}. In the LLM reasoning domain, recent methods apply test-time RL using self-generated signals: TTRL~\cite{zuo2025ttrl} uses majority voting for pseudo-labels, Intuitor~\cite{zhao2025learning} and EMPO~\cite{zhang2025right} use entropy-based rewards. These methods share a structural deficiency: they perform only policy refinement while neglecting reward calibration, which causes the reward signal to drift from true correctness as the policy evolves. \ours addresses this by interleaving critic recalibration (E-step) with policy optimization (M-step), maintaining a tight ELBO and enabling sustained improvement beyond the ceilings of prior methods.

\section{Method}\label{sec:method}
We propose Test-time Expectation-Maximization Policy Optimization (\ours), a TTT framework that initializes actor and critic via RLVR on labeled data $D_L$, then continuously improves on unlabeled test questions by alternating between critic calibration and policy refinement following the EM algorithm. This section is organized as follows: we first formalize the problem setup (Section~\ref{sec:problem_setup}), then derive an EM-inspired variational lower bound as optimization objective (Section~\ref{sec:elbo}), and finally detail the alternating E-step critic calibration (Section~\ref{sec:estep}) and M-step policy optimization (Section~\ref{sec:mstep}).

\subsection{Problem Setup}\label{sec:problem_setup}

We consider a setting where the model has access to a labeled dataset $D_L$ containing ground-truth answers and a set of unlabeled test questions $D_u$ where the correct responses are unknown. Our goal is to enable LRMs to continuously self-improve during the test phase. Formally, we aim to maximize the expected log-probability of generating a correct response given a question $x$. Let $\theta$ denote the parameters of the LRM, and $P(\text{Correct}|x; \theta)$ represent the probability that the model produces a correct output for a given input $x$. The global objective function is defined as follows:
\begin{equation}
J(\theta) = \mathbb{E}_x [\log P(\text{Correct}|x; \theta)],
\end{equation}
where the marginal probability $P(\text{Correct}|x; \theta)$ is obtained by marginalizing over all possible generated responses $y$ as
\begin{equation} 
P(\text{Correct}|x; \theta) = \sum_y P(\text{Correct}|x, y) \pi_{\theta}(y|x).
\end{equation}
Here, $\pi_\theta(y|x)$ denotes the policy, which represents the output distribution of the LRMs.

\subsection{Variational Lower Bound Objective}\label{sec:elbo}
We first derive the evidence lower bound (ELBO) that enables optimization when ground-truth correctness is unobserved, showing how the EM framework decomposes the objective into an estimable lower bound and a KL divergence term.

The fundamental challenge in test-time training is that the response correctness for $x\in D_u$ is unobserved. In such scenarios, the optimal response distribution $P(y|x, \text{Correct})$ is a latent variable. To optimize $J(\theta)$ under these conditions, we employ the Expectation-Maximization (EM) framework. By introducing an auxiliary distribution $q(y|x)$, we derive the Evidence Lower Bound (ELBO):
\begin{align}
J(\theta) &= \sum_{x \in D_u} \log \sum_{y} q(y|x) \frac{P(\text{Correct}|y, x) \pi_{\theta}(y|x)}{q(y|x)} \
\\
&= \sum_{x \in D_u} \left( \sum_{y} q(y|x) \log \frac{P(\text{Correct}|y, x) \pi_{\theta}(y|x)}{q(y|x)} + KL(q(y|x) || P(y|x, \text{Correct})) \right).
\end{align}
Intuitively, this decomposition says: maximizing the lower bound corresponds to (i) increasing the expected log-likelihood of responses that are likely to be correct, and (ii) bringing the auxiliary distribution $q$ closer to the true posterior $P(y|x, \text{Correct})$.

By omitting the non-negative KL divergence term, we obtain the objective $\mathcal{L}(q, \theta)$, where $J(\theta) \ge \mathcal{L}(q, \theta)$. The equality holds if and only if $q(y|x)$ perfectly matches the posterior distribution $P(y|x, \text{Correct})$. This allows for iterative refinement by alternating between estimating the distribution of correct responses and maximizing model likelihood.

\subsection{Expectation Step: Posterior Estimation via Critic}\label{sec:estep}

Then we describe how to train a critic model on labeled data to approximate the posterior distribution over correct responses, thereby grounding the reward signal in external supervision. In the E-step, we keep the current policy $\pi_{\theta_0}$ fixed and seek the optimal auxiliary distribution $q^*(y|x)$ that maximizes the lower bound. This optimal distribution corresponds to the posterior probability of the response conditioned on its correctness:
\begin{equation}
q(y|x) = P(y|x, \text{Correct}) = \frac{P(\text{Correct}|y, x) \pi_{\theta_0}(y|x)}{P(\text{Correct}|x)}.
\end{equation} To approximate the unknown term $P(\text{Correct}|y, x)$, we train a critic model $V_{\phi}(x, y_t) \in \mathbb{R}$ using the labeled data $D_L$, where $t\geq 0$ is the token index. The critic is optimized by minimizing the MSE between its token-level predictions and the ground-truth outcomes. For each response $y$ associated with a query $x \in D_L$, the critic is trained to perform token-level value estimation. Formally, the critic parameters $\phi$ are updated by
\begin{equation} \label{eq:critic}
\mathcal{L}_{\text{critic}}(\phi) = \mathbb{E}_{(x, y, \mathcal{I}) \sim D_L} \| V_{\phi}(x, y_t) - \mathcal{I} \|_2^2,
\end{equation}
where $\mathcal{I} \in \{0, 1\}$ denotes the binary correctness indicator. This training regime ensures the critic serves as a reliable proxy for the expected correctness of generated response $y$.

Once optimized, the critic provides a tractable surrogate for the posterior distribution. Since the critic is trained to predict outcome correctness, its last-token value $V_\phi(x, y_T)$ directly reflects the likelihood of a correct response, enabling the optimal auxiliary distribution to be approximated by reweighting the model's current outputs with the critic scores as follows:
\begin{equation} 
q(y|x) \propto V_{\phi}(x, y_T) \pi_{\theta_0}(y|x),
\end{equation}
where $T$ is the response length of $y$. This step effectively identifies high-quality responses from the model's own generations to serve as a surrogate for the missing ground-truth labels.

\subsection{Maximization Step: Policy Optimization}\label{sec:mstep}

Finally, we formulate the policy update as a weighted maximum likelihood estimation using critic-derived rewards, and implement it via a policy gradient RL framework with token-level advantage estimation. In the M-step, we fix the auxiliary distribution $q(y|x)$ and update the model parameters $\theta$ by maximizing the lower bound $\mathcal{L}(q, \theta)$. The optimization problem is formulated as follows:
\begin{equation}
\theta_{\text{new}} = \arg \max_{\theta} \sum_{x \in D_u} \sum_{y} q(y|x) \log \left( P(\text{Correct}|y, x) \pi_{\theta}(y|x) \right).
\end{equation}
By removing terms independent of $\theta$, the objective simplifies to a weighted maximum likelihood estimation. Substituting the approximation of $q^*(y|x)$ derived in the E-step, the update rule becomes
\begin{equation}
\theta_{\text{new}} = \arg \max_{\theta} \sum_{x \in D_u} \sum_{y} V_{\phi}(x, y_T) \pi_{\theta_0}(y|x) \log \pi_{\theta}(y|x).
\end{equation}
Given that $\pi_{\theta_0}(y|x)$ represents the sampling distribution, the inner summation can be interpreted as an expectation. We further simplify the objective by focusing on the weighted log-likelihood of the sampled responses as
\begin{equation}
\theta_{\text{new}} = \arg \max_{\theta} \sum_{x \in D_u} \sum_{y} V_{\phi}(x, y_T) \log \pi_{\theta}(y|x).
\end{equation}
This optimization is then solved using policy gradient methods, facilitating the continuous self-refinement of the policy model based on its reasoning trajectory on the unlabeled open questions.

\begin{figure}[H]
\centering
    \includegraphics[width=0.8\textwidth]{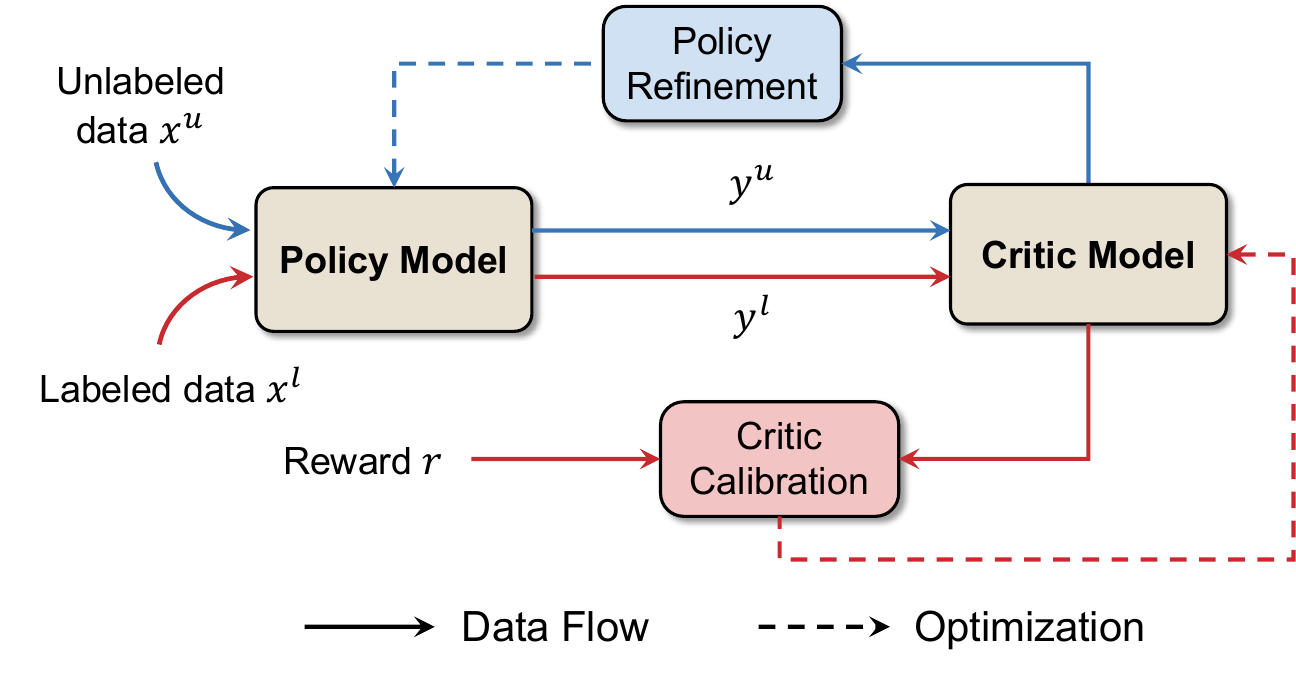}
    \caption{\ours alternates between (i) \textit{Critic Recalibration (E-step)}: the critic is periodically updated using verifiable rewards from $D_L$ to maintain a grounded and informative reward signal, and (ii) \textit{Policy Refinement (M-step)}: the actor generates reasoning trajectories on unlabeled questions $D_u$ and optimizes against critic-derived rewards. This alternating EM-style procedure enables sustained self-improvement beyond the RLVR plateau.}
    \label{fig:pipeline}
\end{figure}

To effectively optimize this objective while ensuring variance reduction, we implement the M-step update via a policy gradient-based RL framework. In this setting, the value prediction of the critic at the terminal token of the sequence is treated as the ground-truth external reward $R = V_{\phi}(x, y)$ for the entire response. To derive a stable training signal, we utilize the critic's intermediate value predictions $V_{\phi}(x, y_{1:t})$ as token-varying baselines $b_t$. The advantage $A_t$ for each token $y_t$ is defined as the discrepancy between the final realized reward and the expected value at step $t$:
\begin{equation} \label{eq:advantage}
A_t = R - V_{\phi}(x, y_{1:t}).
\end{equation}
This formulation ensures that tokens contributing to a higher last-token value prediction receive a positive reinforcement signal, while those leading to deviations from the predicted value are penalized. The policy optimization objective for the M-step is
\begin{equation} \label{eq:policy}
\mathcal{L}_{\text{policy}}(\theta) = -\mathbb{E}_{x \in D_u, y \sim \pi_{\theta}}\left[ \sum_{t=1}^T A_t \log \pi_\theta(y_t | x, y_{<t}) \right].
\end{equation}
By alternating between the critic recalibration (E-step) and the policy refinement (M-step), the model achieves continuous, scalable self-improvement on unlabeled open reasoning problems. The full alternating procedure is summarized in Figure~\ref{fig:pipeline} and Algorithm~\ref{alg:stpo}.

\begin{algorithm}[H]
\caption{Test-time Expectation-Maximization Policy Optimization (\ours)}
\label{alg:stpo}
\SetAlgoLined
\KwIn{Labeled dataset $D_L$, unlabeled test set $D_u$, initial policy $\pi_{\theta_0}$, critic $V_{\phi_0}$, total iterations $N$}
\KwOut{Updated policy $\pi_{\theta_N}$}
\tcc{Stage 1: Initialization}
Train actor $\pi_{\theta_0}$ and critic $V_{\phi_0}$ via RLVR on $D_L$\;
\For{$k = 1$ \KwTo $N$}{
  \tcc{Stage 2: Alternating test-time training}
    \tcc{E-step: Critic recalibration}
    Sample $(x, y, \mathcal{I})$ from $D_L$\;
    Update critic $\phi$ by minimizing $\mathcal{L}_{\text{critic}}(\phi)$ (Eq.~\eqref{eq:critic})\;
  \tcc{M-step: Policy refinement}
  Sample $x \sim D_u$, generate responses $y \sim \pi_{\theta}$\;
  Compute advantages $A_t = V_\phi(x, y_T) - V_\phi(x, y_{1:t})$ (Eq.~\eqref{eq:advantage})\;
  Update policy $\theta$ by minimizing $\mathcal{L}_{\text{policy}}(\theta)$ (Eq.~\eqref{eq:policy})\;
    }
\Return $\pi_{\theta_N}$\;
\end{algorithm}

\section{Experiments}\label{sec:experiments}
We empirically validate \ours across four dimensions. We first describe the experimental setup (Section~\ref{sec:exp_setup}), then demonstrate sustained scalability beyond RLVR ceilings (Section~\ref{sec:scalability}), show that \ours preserves output diversity while baselines collapse (Section~\ref{sec:diversity}), validate generalization to non-math reasoning tasks (Section~\ref{sec:versatility}), and finally ablate the alternating training design to confirm the necessity of each component (Section~\ref{sec:ablation}).

\subsection{Experimental Setup}\label{sec:exp_setup}
We evaluate \ours on both mathematical reasoning and general domain reasoning tasks, covering a diverse set of base models, datasets, and baselines.
\begin{itemize}
    \item For math experiments, we select Qwen3~\cite{qwen3} and OLMO3~\cite{olmo3} as the base models. We first initialize the actor and critic models with PPO on DAPO-Math-17K~\cite{yu2025dapo} as the labeled training dataset, using standard RLVR to establish a starting point. Subsequently, we perform test-time training driven by \ours on a test set consisting of AIME 2024, AIME 2025, and Beyond AIME~\cite{bytedance_seed_2025_beyondaime}. For a more comprehensive evaluation, we introduce AIME 2026 and OlymMath~\cite{zhang2025olymmath} as holdout test sets to assess generalization. For the RL training recipes, we set the batch size to 256 and the mini-batch size to 64 for models up to 8B. A batch size of 128 and a mini-batch size of 32 for the 14B model due to GPU memory constraints. The maximum response length is 16K. To stabilize off-policy training, we implement the sequence clip mechanism with a dual-clip ratio of $3 \times 10^{-4}$ and $5 \times 10^{-4}$ as suggested by GSPO~\cite{zheng2025group}. We compare \ours against several representative baselines, including standard RLVR trained via PPO and representative self-training methods TTRL~\cite{zuo2025ttrl} and EMPO~\cite{zhang2025right}. We report \textit{avg@16} accuracy (average accuracy over 16 independent samples per problem) and \textit{pass@8} (the fraction of problems solved by at least one of 8 samples).
    
    \item For general domain reasoning tasks, we initialize the actor and critic via PPO on Dolci-RL-Zero-General~\cite{olmo3}, a 12.8K labeled corpus. Subsequently, we perform test-time training driven by \ours{} on a test set consisting of BigBenchHard~\cite{suzgun2023challenging}, AGI Eval~\cite{zhong2024agieval}, and ZebraLogic~\cite{lin2025zebralogic}. For a more comprehensive evaluation, we introduce GPQA-Diamond~\cite{rein2024gpqa} as a holdout test set to assess generalization. All RL training hyperparameters remain identical to the math experiments. To ensure high-fidelity assessment across these diverse domains, we employ \texttt{gpt-oss-120b}~\cite{agarwal2025gpt} as the judge model for correctness verification. Since the volume of BBH, AGI Eval, and ZebraLogic is already sufficient to ensure stable evaluation, we solely report \textit{Avg@1} accuracy for them, while reporting \textit{Avg@8} and \textit{Pass@8} for the highly complex GPQA-Diamond benchmark.
\end{itemize}

\subsection{Scalability (RQ1)}\label{sec:scalability}

We first evaluate whether \ours can sustainably improve model performance beyond the RLVR training ceiling by leveraging unlabeled test data. We compare \ours against the RLVR baseline and two representative self-rewarding TTT methods (TTRL and EMPO) across three model families and five benchmarks. Results are shown in Table~\ref{tab:math-main}.

\ours \textbf{significantly outperforms all baselines across model scales and benchmarks.} For instance, on the AIME 24 dataset, \ours improves the \textit{avg@16} accuracy of OLMO3-7B-Base from 33.0\% to 51.1\%, and Qwen3-14B-Base from 42.3\% to 65.8\%. Gains are particularly pronounced on the most challenging benchmarks (AIME 24 and AIME 25), where prior methods show the largest degradation relative to \ours. This scalability stems from the EM-based alternating structure. By periodically recalibrating the critic on labeled data, \ours prevents the reward signal from drifting as the policy evolves, which is the failure mode that causes prior methods to plateau once the model becomes overconfident in a narrow set of reasoning patterns. In contrast, the grounded critic in \ours continues to provide informative gradients, enabling the model to explore high-quality reasoning paths for challenging questions at test-time.

We further evaluate \ours's ability to surpass the RLVR performance upper bound by continuing from a converged OLMO3 model (pre-trained for 192 steps on DAPO-Math-17K). As shown in Figure~\ref{fig:ablation_ppo}, further RLVR training with PPO yields negligible gains, whereas \ours-based test-time training produces a consistent performance surge over 200 iterations. The widening gap between these two curves confirms that \ours translates additional test-time compute into measurable capability gains, transcending the limits of standard RLVR.

\begin{table}[ht]
\centering
\caption{\textbf{Main results on mathematical reasoning benchmarks.} We report avg@16 accuracy and pass@8 over 16 independent samples across five benchmarks and the absolute Improvement via TTT of \ours over the Zero-RL baseline. \label{tab:math-main}}
\small 
\setlength{\tabcolsep}{2.5pt} 
\begin{tabular*}{\linewidth}{@{\extracolsep{\fill}} 
    l 
    >{\columncolor{gray!10}}c >{\columncolor{gray!10}}c  
    >{\columncolor{gray!10}}c >{\columncolor{gray!10}}c  
    >{\columncolor{gray!10}}c >{\columncolor{gray!10}}c  
    >{\columncolor{blue!5}}c  >{\columncolor{blue!5}}c  
    >{\columncolor{blue!5}}c  >{\columncolor{blue!5}}c   
}
\toprule
\multirow{2}{*}{\textbf{Method}} 
    & \multicolumn{2}{c}{\textbf{Beyond AIME}} 
    & \multicolumn{2}{c}{\textbf{AIME 24}} 
    & \multicolumn{2}{c}{\textbf{AIME 25}} 
    & \multicolumn{2}{c}{\textbf{AIME 26}} 
    & \multicolumn{2}{c}{\textbf{OlymMath}} \\
\cmidrule(lr){2-3} \cmidrule(lr){4-5} \cmidrule(lr){6-7} \cmidrule(lr){8-9} \cmidrule(lr){10-11}
& \textbf{Acc} & \textbf{Pass@K} 
& \textbf{Acc} & \textbf{Pass@K} 
& \textbf{Acc} & \textbf{Pass@K} 
& \textbf{Acc} & \textbf{Pass@K} 
& \textbf{Acc} & \textbf{Pass@K} \\
\midrule
\multicolumn{11}{l}{\textit{Frontier Models}} \\
Oat-Zero-7B       & 9.4    & 19.4    & 30.2 & 46.1    & 12.3 & 33.7    & 16.7    & 26.3    & 11.1    & 22.6    \\
MiMo-Zero-RL-7B            & 14.6    & 33.1    & 37.7 & 63.9    & 32.3 & 51.9    & 35.0    & 52.8    & 16.7    & 36.1    \\
OLMO3.1-Zero-RL-7B        & 13.8 & 32.1    & 31.9 & 56.3    & 26.5 & 39.7    & 24.0 & 42.4    & 14.3 & 42.3\\
\midrule
\textbf{OLMO3-7B-Base}     &      &      &      &      &      &      &      &      &      &      \\
\quad {\color{gray}$\hookrightarrow$} Zero-RL (PPO) & 17.6 & 38.8 & 33.0 & 56.1 & 26.3 & 41.1 & 26.7 & 42.8 & 18.9 & 43.3 \\
\quad\quad {\color{gray}$\hookrightarrow$} TTRL     & 21.8 & 22.3 & 40.8 & 45.6 & 27.1 & 30.7 & 22.8 & 39.2 & 18.9 & 33.0 \\
\quad\quad {\color{gray}$\hookrightarrow$} EMPO     & 21.3 & 28.4 & 41.6 & 43.3 & 26.7 & 29.5 & 23.6  & 39.7  & 18.7 & 32.9 \\
\quad\quad {\color{gray}$\hookrightarrow$} \ours    & 24.5 & 44.0 & 51.1 & 61.6 & 37.0 & 52.5 & 30.1 & 49.4 & 23.5 & 51.6 \\
\rowcolor{red!5} \quad\quad \textbf{Improvement via TTT} & \textbf{+6.9} & \textbf{+5.2} & \textbf{+18.1} & \textbf{+5.5} & \textbf{+10.7} & \textbf{+11.4} & \textbf{+3.4} & \textbf{+6.6} & \textbf{+4.6} & \textbf{+8.3} \\
\addlinespace[0.5em] 
\textbf{Qwen3-8B-Base}     &      &      &      &      &      &      &      &      &      &      \\
\quad {\color{gray}$\hookrightarrow$} Zero-RL (PPO) & 15.6 & 33.6 & 26.3 & 53.0 & 25.4 & 44.8 & 21.9 & 43.7 & 15.0 & 39.9 \\
\quad\quad {\color{gray}$\hookrightarrow$} TTRL     & 18.7 & 20.0 & 29.0 & 30.0 & 32.8 & 33.3 & 13.8 & 25.0 & 11.4 & 25.3 \\
\quad\quad {\color{gray}$\hookrightarrow$} EMPO     & 16.7 & 23.3 & 32.3 & 26.7 & 33.3 & 35.4 & 19.4 & 33.3 & 13.1 & 26.7 \\
\quad\quad {\color{gray}$\hookrightarrow$} \ours    & 20.0 & 36.7 & 42.7 & 61.1 & 40.8 & 60.4 & 24.2 & 50.7 & 18.7 & 43.3 \\
 \rowcolor{red!5} \quad\quad \textbf{Improvement via TTT} & \textbf{+4.4} & \textbf{+3.1} & \textbf{+16.4} & \textbf{+8.1} & \textbf{+15.4} & \textbf{+15.6} & \textbf{+2.3} & \textbf{+7.0} & \textbf{+3.7} & \textbf{+3.4} \\
\addlinespace[0.5em] 
\textbf{Qwen3-14B-Base}    &      &      &      &      &      &      &      &      &      &      \\
\quad {\color{gray}$\hookrightarrow$} Zero-RL (PPO) & 24.9 & 50.0 & 42.3 & 69.1 & 37.1 & 59.0 & 38.1 & 70.0 & 24.2 & 51.6 \\
\quad\quad {\color{gray}$\hookrightarrow$} TTRL     & 25.5 & 29.4  & 53.1 & 56.7 & 40.8 & 45.8 & 31.7 & 43.0 & 18.3 & 31.7 \\
\quad\quad {\color{gray}$\hookrightarrow$} EMPO     & 27.6 &  31.4 & 55.6 & 59.7 & 44.6 & 46.7 & 28.3 & 49.5 & 17.7 & 31.9 \\
\quad\quad {\color{gray}$\hookrightarrow$} \ours    & 29.3   & 46.3 & 65.8   & 73.3 & 44.6   & 60.0 & 38.8   & 70.0 & 25.8   & 50.2 \\
\rowcolor{red!5} \quad\quad \textbf{Improvement via TTT} & \textbf{+4.4} & -3.7 & \textbf{+23.5} & \textbf{+4.2} & \textbf{+7.5} & \textbf{+1.0} & \textbf{+0.7} & \textbf{0.0} & \textbf{+1.6} & -1.4 \\
\bottomrule
\end{tabular*}
\end{table}

\subsection{Diversity (RQ2)}\label{sec:diversity}
\paragraph{\ours improves model performance without compromising diversity.} A critical yet often overlooked aspect of test-time training is the preservation of output diversity. While many methods achieve short-term \textit{avg@16} accuracy gains, they frequently suffer from diversity collapse, i.e., the model converges to a narrow set of reasoning patterns, causing pass@k to degrade even as mean@k improves. This phenomenon fundamentally limits the scalability of test-time training, as additional samples yield diminishing returns. As reported in Table~\ref{tab:math-main}, \ours consistently maintains high pass@k scores across all benchmarks, while representative baselines exhibit significant diversity degradation. For instance, on the Qwen3-14B-Base model, \ours achieving a pass@k of 73.0 on AIME 24 and 64.3 on AIME 25, substantially outperforming TTRL's 56.7 and 43.3, respectively. Similarly, EMPO records 60.0 on Beyond AIME and 46.7 on AIME 25, trailing \ours by 7.6 and 17.6 points.

This performance gap stems from fundamental differences in how each method constructs its self-training signal. EMPO and TTRL rely on entropy or self-consistency to encourage consensus, which inherently favors the most common reasoning path regardless of its quality. As training progresses, the model becomes increasingly confident in its dominant mode, suppressing alternative valid solutions and causing the output distribution to collapse. In contrast, \ours employs a dynamically calibrated critic that assigns continuous, quality-aware scores to each generated response. This mechanism naturally preserves diversity among high-quality solutions while down-weighting incorrect but frequently generated patterns.

\begin{figure}[ht]
    \centering
    \begin{minipage}[t]{0.48\textwidth}
        \centering
        \includegraphics[width=\textwidth]{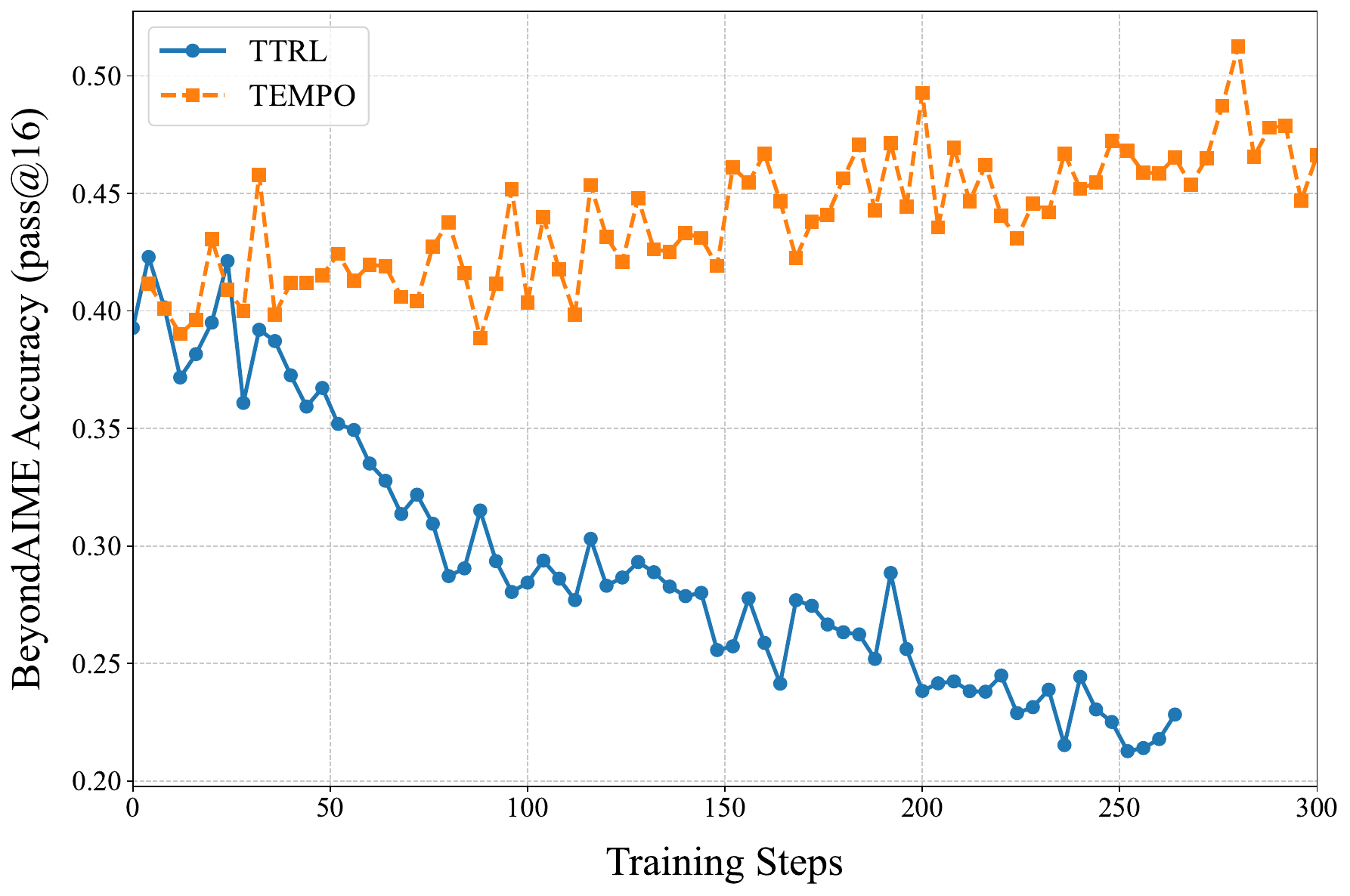}
        \caption{\textbf{TEMPO preserves model diversity.} We compare \ours with TTRL on Beyond AIME pass@16. While TTRL consistently degrades pass@16 throughout training, \ours maintains and steadily improves pass@16. This reveals a fundamental distinction: prior methods trade away exploration 
  capacity for short-term performance, whereas \ours sustains genuine reasoning diversity as a foundation for continued self-improvement.}
        \label{fig:diversity}
    \end{minipage}
    \hfill
    \begin{minipage}[t]{0.48\textwidth}
        \centering
        \includegraphics[width=\textwidth]{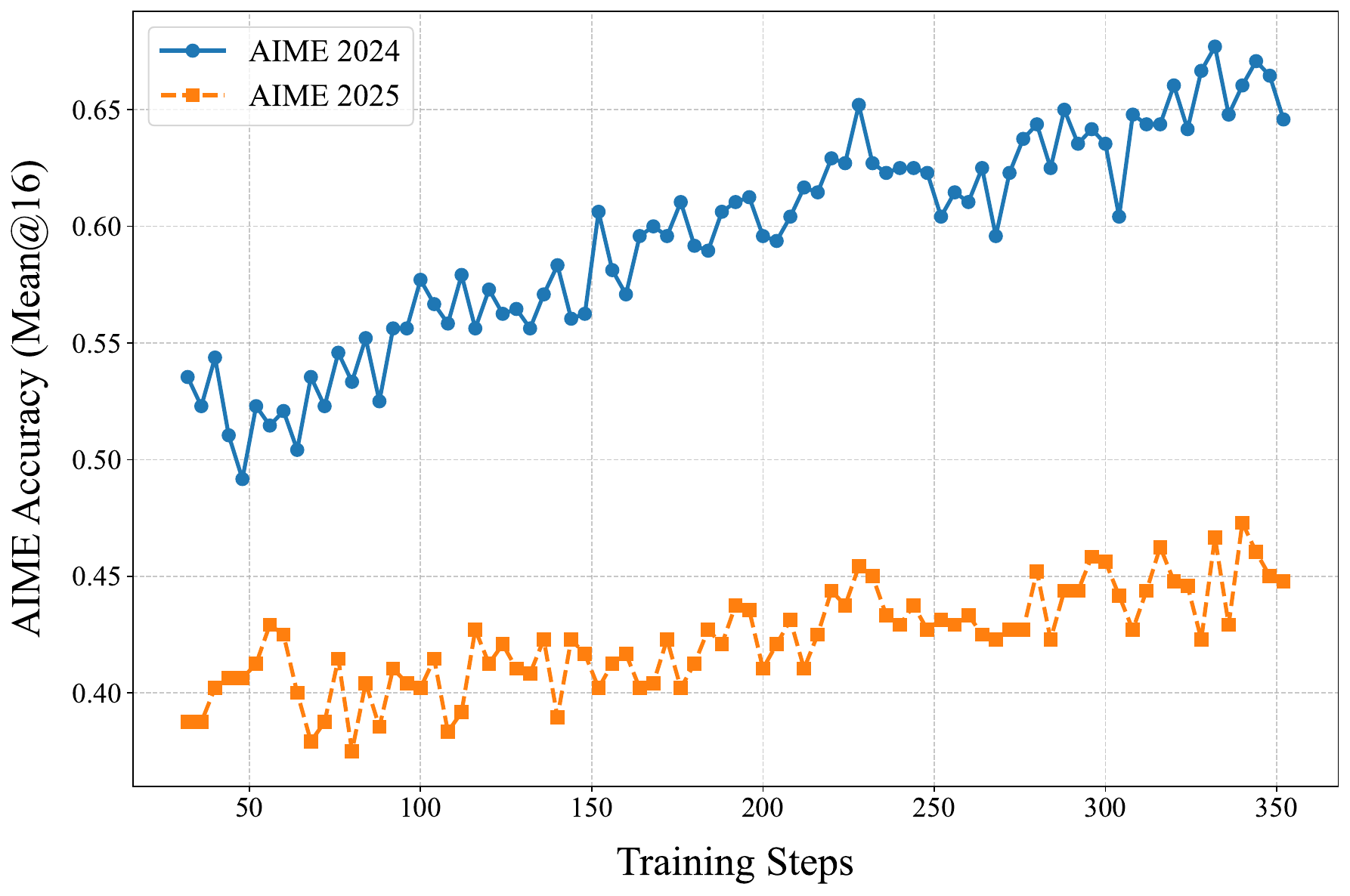}
        \caption{\textbf{TEMPO continues to improve beyond reported results.} The numbers reported in our main table correspond to \ours    
  trained on Qwen3-14B for 224 steps. As shown here, performance on both AIME 2024 and AIME 2025 has not plateaued at that                    
  checkpoint. The \textit{avg@16} accuracy continues to rise with further training. This suggests that the results we report are conservative, and \ours has the 
  potential for even more gains given additional compute.}
        \label{fig:scaling}
    \end{minipage}
\end{figure}

\subsection{Versatility (RQ3)} \label{sec:versatility} 

\paragraph{\ours is applicable for general reasoning tasks beyond math.}
We investigate whether the effectiveness of \ours{} extends beyond mathematical problem-solving. To this end, we evaluate both the OLMO3-7B-Base and Qwen3-8B-Base models across four diverse, general-reasoning domains: BigBenchHard (BBH), AGI Eval, ZebraLogic, and the expert-level GPQA-Diamond. As shown in Table~\ref{tab:general}, \ours demonstrates significant versatility and robust performance across both OLMO3 and Qwen3 families.

Specifically, \ours achieves substantial absolute gains of +21.4 on BBH and +24.5 on AGI Eval for OLMO3-7B. These enhancements enable the 7B model to surpass specialized frontier models like General-Reasoner-7B and MiMo-Zero-RL-7B. On the Qwen3-8B model, despite its much higher starting performance, \ours still yields consistent improvements, particularly on the logic-intensive ZebraLogic (\textbf{+8.2}) and the expert-level GPQA-Diamond (\textbf{+5.0} Avg@8). This suggests that \ours is not merely overfitting to specific reasoning patterns but is effectively enhancing the underlying logical capabilities across different base model architectures.

When compared to other test-time training methods, \ours demonstrates a more balanced and reliable performance profile. Prior self-training methods often exhibit systemic sensitivity to the initial policy's capability or the domain's complexity. In contrast, \ours achieves massive gains on OLMO3 across the board, outperforming TTRL by over 20 points on BBH and AGI Eval. Furthermore, on the Qwen3 model, \ours maintains an edge in AGI Eval and ZebraLogic. These results indicate that the alternating training design in \ours is particularly robust when the initial policy is less mature, providing a grounded signal that prevents the performance stagnation or collapse observed in prior self-training baselines.

\begin{table}[ht]
\centering
\caption{\textbf{Generalization to reasoning tasks beyond math-only domains.} We report Avg@1 for BigBenchHard, AGI Eval, and ZebraLogic, alongside Avg@8 and Pass@8 over 8 independent samples for GPQA-Diamond. We also highlight the absolute \textit{Improvement via TTT} of \ours over the Zero-RL baseline across diverse domains. \label{tab:general}}
\small
\setlength{\tabcolsep}{3pt}
\begin{tabular*}{\linewidth}{@{\extracolsep{\fill}} 
    l 
    >{\columncolor{gray!6}}c
    >{\columncolor{gray!6}}c 
    >{\columncolor{gray!6}}c 
    >{\columncolor{blue!5}}c
    >{\columncolor{blue!5}}c
}
\toprule
\textbf{Method} 
    & \textbf{BBH}
    & \textbf{AGI} 
    & \textbf{Zebra} 
    & \multicolumn{2}{c}{\textbf{GPQA-Diamond}} \\
\cmidrule(lr){5-6}
    & (Avg@1) & (Avg@1) & (Avg@1) & (Avg@8) & (Pass@8) \\
\midrule
\textit{Frontier Models} & & & & & \\
Olmo-3-7B-RL-Zero-General    & 56.5 & 51.9 & 25.7 & 28.9 & 69.0 \\
MiMo-Zero-RL-7B              & 61.4 & 53.6 & 30.3 & 18.8 & 45.8 \\
General-Reasoner-7B          & 65.6 & 63.6 & 25.9 & 35.1 & 68.6 \\
General-Reasoner-14B         & 78.2 & 73.4 & 44.5 & 44.4 & 70.3 \\
\midrule
\textbf{OLMO3-7B-Base} & & & & & \\
\quad {\color{gray}$\hookrightarrow$} Zero-RL (PPO) & 46.8 & 37.9 & 22.2 & 21.9 & 62.1 \\
\quad\quad {\color{gray}$\hookrightarrow$} TTRL     & 45.4 & 38.2 & 22.2 & 28.5 & 67.6 \\
\quad\quad {\color{gray}$\hookrightarrow$} EMPO     & 52.9 & 50.2 & 23.5 & 27.7 & 61.6 \\
\quad\quad {\color{gray}$\hookrightarrow$} \ours    & 68.2 & 62.4 & 35.1 & 32.4 & 69.4 \\
\rowcolor{red!5}
\quad\quad {\textbf{Improvement via TTT}} 
    & {\textbf{+21.4}} 
    & {\textbf{+24.5}} 
    & {\textbf{+12.9}} 
    & {\textbf{+10.5}} 
    & {\textbf{+7.3}} \\
\midrule
\textbf{Qwen3-8B-Base} & & & & & \\
\quad {\color{gray}$\hookrightarrow$} Zero-RL (PPO) & 69.9 & 65.7 & 25.7 & 32.2 & 62.4 \\
\quad\quad {\color{gray}$\hookrightarrow$} TTRL     & 74.9 & 68.5 & 31.7 & 41.1 & 73.0 \\
\quad\quad {\color{gray}$\hookrightarrow$} EMPO     & 66.7 & 65.1 & 26.3 & 39.8 & 70.8 \\
\quad\quad {\color{gray}$\hookrightarrow$} \ours    & 74.2 & 70.1 & 33.9 & 37.2 & 65.3 \\
\rowcolor{red!5}
\quad\quad {\textbf{Improvement via TTT}} 
    & {\textbf{+4.3}} 
    & {\textbf{+4.4}} 
    & {\textbf{+8.2}} 
    & {\textbf{+5.0}} 
    & {\textbf{+2.9}} \\
\bottomrule
\end{tabular*}
\vspace{-0.5em}
\end{table}

\subsection{Ablation (RQ4)}\label{sec:ablation}

We conduct two ablation studies to isolate the contributions of key design choices in \ours:
(1) \textit{Frozen critic}: the critic is trained once on $D_L$ and kept fixed throughout all policy updates, removing the E-step recalibration;
(2) \textit{Supervised continuation}: the model continues training on the labeled dataset $D_L$ using standard PPO without any test-time updates on unlabeled data.

\begin{figure}[ht]
    \centering
    \begin{minipage}[t]{0.48\textwidth}
        \centering
        \includegraphics[width=\textwidth]{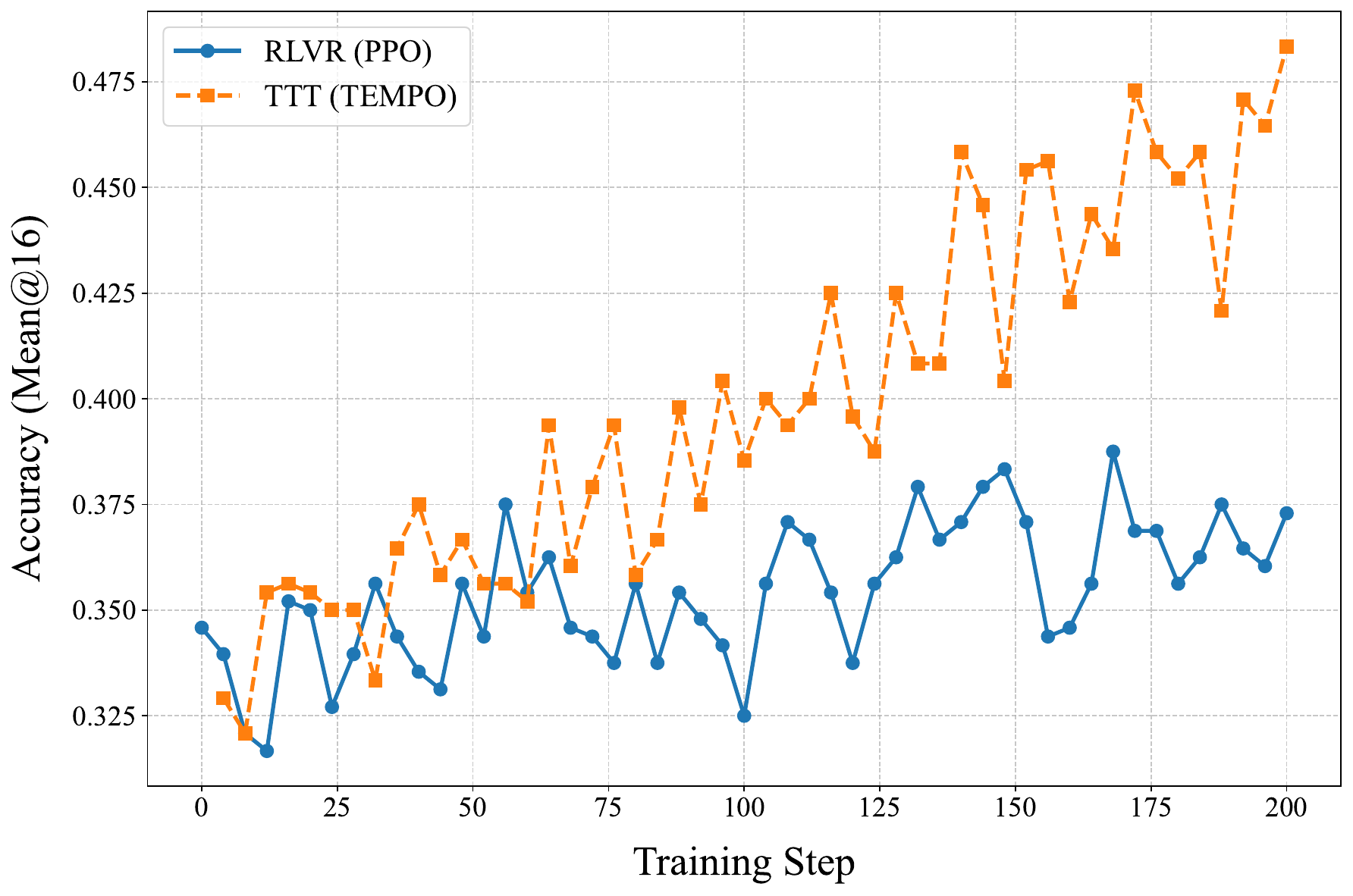}
        \caption{\textbf{The Superiority of Test-time training.} Starting from a converged OLMO3 model (192 PPO steps on DAPO-Math-17K), we compare continuing supervised PPO on the same labeled data (blue) with \ours test-time training on unlabeled questions (orange). Supervised PPO saturates immediately with negligible gains, while \ours achieves a steady 15+ point \textit{avg@16} accuracy improvement over 200 iterations, confirming that test-time training on novel problems pushes the performance boundary.}
        \label{fig:ablation_ppo}
    \end{minipage}
    \hfill
    \begin{minipage}[t]{0.48\textwidth}
        \centering
        \includegraphics[width=\textwidth]{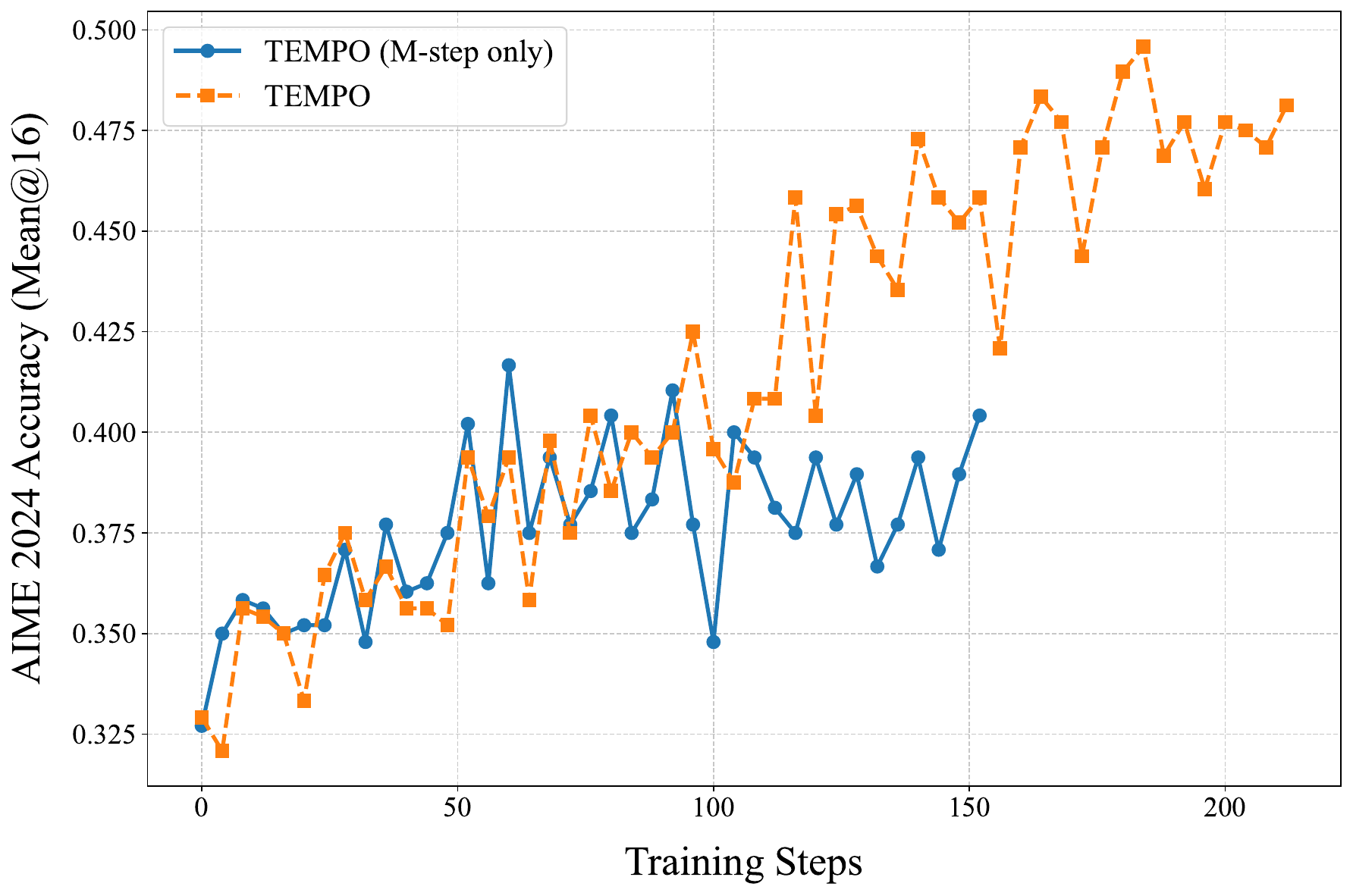}
        \caption{\textbf{Necessity of alternating critic recalibration.} We compare the full \ours (orange) with a frozen-critic variant (blue) where the critic is trained once on $D_L$ and never updated. The frozen critic initially matches \ours but plateaus after \textasciitilde100 iterations as it becomes misaligned with the evolving policy, while the full model continues to improve. This confirms that periodic E-step recalibration is essential for sustained self-improvement.}
        \label{fig:ablation_em}
    \end{minipage}
\end{figure}

As shown in Figure~\ref{fig:ablation_ppo}, the supervised-only continuation (supervised PPO) quickly saturates and yields negligible gains over 200 additional training steps, confirming that the model has already converged on the labeled distribution. In stark contrast, \ours exhibits a steady and consistent upward trajectory from the same starting point. The widening gap between these two curves grows from near-zero at step 0 to over 15 \textit{avg@16} accuracy points by step 200. This gap represents performance gains that are entirely attributable to test-time training on unlabeled open questions. This result demonstrates that once a model has converged on its supervised training data, further optimization on the same distribution cannot unlock additional capability. Only exposure to novel, challenging test-time problems can push the model beyond its established boundaries.

Figure~\ref{fig:ablation_em} reveals the critical role of the alternating training design. The frozen-critic variant initially matches the performance of the full \ours, confirming that a well-calibrated critic can provide useful signals in the early stages. However, as training progresses, its improvement curve flattens and eventually plateaus, diverging sharply from the full model's sustained growth. This degradation occurs because a static critic gradually becomes misaligned with the evolving policy: as the actor generates increasingly sophisticated reasoning paths, the frozen critic---trained on an earlier, less capable policy's outputs---fails to accurately evaluate these new patterns. The resulting mismatch between the critic's scores and the true correctness of responses introduces noise into the policy gradient, ultimately stalling further improvement. This ablation validates that the E-step critic recalibration is not a mere implementation detail but a \textit{necessary condition} for sustained self-improvement: without periodic grounding on labeled data, the critic's evaluations drift, and the entire self-training loop collapses.

\section{Discussion}\label{sec:discussion}
This section interprets representative TTT methods through the lens of our EM framework, showing how TTRL and EMPO reduce to degenerate cases that omit the E-step, and explaining why our dynamically calibrated critic avoids the self-reinforcement trap.

\paragraph{A unified perspective on LRM test-time training.} The proposed \ours framework offers a principled Expectation-Maximization (EM) interpretation of test-time training for LRMs. By iteratively alternating between posterior estimation (E-step) and policy optimization (M-step), \ours ensures that the Evidence Lower Bound (ELBO) remains a tight surrogate for the true objective $J(\theta)$, thereby preventing the optimization from diverging as the model self-improves. This theoretical lens reveals a critical insight: several representative test-time training methods, including EMPO and TTRL, can be understood as heuristic and degenerate instances of the EM algorithm. Specifically, these methods effectively execute only the M-step using self-generated pseudo-labels for policy updates while entirely neglecting the E-step that should calibrate the quality of those labels.

To make this connection concrete, we reconsider how TTRL constructs its training signal. In TTRL, the auxiliary distribution $q(y|x)$, which in the EM framework should approximate the true posterior $P(y|x, \text{Correct})$, is reduced to a binary indicator based on majority consensus:
\begin{equation}
q(y|x) \propto \mathbbm{1}(y \in \mathcal{Y}_{\rm majority}) \cdot \pi_{\theta_0}(y|x),
\end{equation}
where $\mathbbm{1}(\cdot)$ assigns unit weight to responses that agree with the majority and zero weight to all others. This formulation has two fundamental limitations. First, the consensus set $\mathcal{Y}_{\rm majority}$ is determined solely by the model's current policy. Second, because the training signal is self-generated, it becomes increasingly self-reinforcing: as the model grows more confident in a particular reasoning pattern, that pattern dominates the consensus, which in turn further amplifies the same pattern in subsequent updates. This positive feedback loop is the root cause of the performance plateaus and diversity collapse observed in TTRL.

In contrast, \ours addresses both limitations through a dynamically calibrated critic $V_{\phi}$. Rather than a binary vote, the critic provides a continuous, quality-aware score for each response, enabling fine-grained differentiation among generated samples. More importantly, because the critic is periodically recalibrated on labeled data during the E-step, its evaluations remain grounded in external supervision rather than drifting with the model's own biases. This design ensures that the ELBO stays tight throughout training, allowing \ours to sustain meaningful self-improvement over hundreds of iterations without succumbing to the self-reinforcement trap that limits prior methods.

\section{Limitations}

Despite its advantages, \ours has several limitations that warrant discussion. First, the alternating E/M-step procedure requires maintaining both an actor and a critic model, which increases GPU memory and computational overhead compared to single-model TTT methods such as TTRL. Second, the critic recalibration relies on access to a labeled dataset $D_L$. And the size and distribution of $D_L$ may affect how well the critic generalizes to out-of-domain test questions. Third, our experiments are conducted on math, STEM, and puzzle reasoning tasks; the applicability of \ours to other domains such as code generation remains to be validated. Finally, while the EM perspective provides a principled framing, our theoretical analysis does not include formal convergence guarantees for the alternating optimization, which we leave for future work.

\section{Conclusion}

We present \ours, a scalable test-time training framework for LRMs through an alternating actor-critic optimization. By framing TTT as an EM-style procedure, we identified the missing E-step, i.e., periodic critic recalibration on labeled data as the key deficiency underlying the performance plateaus and diversity collapse of prior baselines. Our theoretical perspective unifies existing TTT approaches as incomplete variants of the EM algorithm, and our empirical results demonstrate that \ours consistently outperforms baselines across model scales and reasoning domains while preserving output diversity. Future work will explore formal convergence guarantees for the alternating procedure, extend the framework to agentic tasks, and investigate the trade-off between frequently-calibrated critic and computational efficiency.

\section{Acknowledgement}
This work is supported by Shanghai Artificial Intelligence Laboratory. The authors thank the P1 team in Shanghai AI Lab for their extensive support of this work, including computational resources, training recipes and insightful discussions.

\bibliographystyle{unsrt}
\bibliography{nips2025}

\end{document}